\documentclass[10pt,twocolumn,letterpaper]{article}

\usepackage[pagenumbers]{cvpr} % To force page numbers, e.g. for an arXiv version

\definecolor{cvprblue}{rgb}{0.21,0.49,0.74}
\usepackage[pagebackref,breaklinks,colorlinks,allcolors=cvprblue]{hyperref}

\usepackage{bm}
\usepackage{amsthm}
\usepackage{amsbsy}
\usepackage{amssymb}
\usepackage{amsmath}
\usepackage{amsfonts}
\usepackage{mathrsfs}
\usepackage{mathtools}

\usepackage{multirow}
\usepackage{makecell}
\usepackage{tabu}
\usepackage{xkeyval}
\usepackage{tabularx}
\usepackage{booktabs}

\usepackage{paralist}  % also compact lists
\usepackage{enumitem}

\usepackage[capitalize,noabbrev]{cleveref}

\theoremstyle{plain}
\newtheorem{theorem}{Theorem}[section]

\theoremstyle{definition}

\theoremstyle{remark}

\usepackage{amsmath}

\def\paperID{15320} % *** Enter the Paper ID here
\def\confName{CVPR}
\def\confYear{2025}

\title{Chanel-Orderer: A Channel-Ordering Predictor for Tri-Channel \\\emph{Natural} Images}

\author{\textbf{Shen Li, Lei Jiang, Wei Wang, Hongwei Hu, Liang Li} \\
\\
\centering Huawei}
\begin{document}
\maketitle
\begin{abstract}
This paper shows a proof-of-concept that, given a typical 3-channel images but in a randomly permuted channel order, a model (termed as Chanel-Orderer) with ad-hoc inductive biases in terms of both architecture and loss functions can accurately predict the channel ordering and knows how to make it right. Specifically, Chanel-Orderer learns to score each of the three channels with the priors of object semantics and uses the resulting scores to predict the channel ordering. This brings up benefits into a typical scenario where an \texttt{RGB} image is often mis-displayed in the \texttt{BGR} format and needs to be corrected into the right order. Furthermore, as a byproduct, the resulting model Chanel-Orderer is able to tell whether a given image is a near-gray-scale image (near-monochromatic) or not (polychromatic). Our research suggests that Chanel-Orderer mimics human visual coloring of our physical natural world.
\end{abstract}    
\section{Introduction}
\label{sec:intro}

The advent of digital imaging has transformed the way we capture, store, and process visual information. However, the reliance on electronic devices and software introduces various challenges, including the correct interpretation of image data. One such challenge is the proper ordering of the color channels in an image, which is critical for accurate representation and subsequent analysis. While the typical representation of color images is in the \texttt{RGB} (Red, Green, Blue) format, various systems and libraries may store images in the \texttt{BGR} (Blue, Green, Red) order, leading to confusion and incorrect display or processing.

In this paper, we present a proof-of-concept that demonstrates the capability of a machine learning model, referred to as Chanel-Orderer, to accurately predict the correct channel order of a given image when the image’s channels are permuted. The model’s architecture and loss functions are designed to incorporate ad-hoc inductive biases that facilitate the learning of color representation of object semantics. As shown in Figure 1, by scoring each of the three channels based on these semantic priors, Chanel-Orderer is able to make accurate predictions about the original channel order. One may notice that the difficulty of this task lies in the ambiguity of image display when the channel order is shuffled: images even ordered in non-\texttt{RGB} format alone may seem valid but still weird; yet, when compared with the valid \texttt{RGB} counterpart, they do not look realistic. Our objective hence is to build a model that is able to overcome this difficulty and learns to restore the valid channel order by predicting the ordering.

An alternative straightforward workaround of this problem is to train a softmax classification model to predict all possible $3!=6$ cases: \texttt{RGB}, \texttt{RBG}, \texttt{GRB}, \texttt{GBR}, \texttt{BRG} and \texttt{BGR}. However, our empirical findings suggests softmax models are inferior to our proposed model. This findings is align with the results from the prior work~\citep{DBLP:journals/corr/abs-1811-12231} which suggests that neural networks may take shortcuts to predict when inductive biases are not sufficiently infused throughout learning. In contrast, our proposed model (termed Chanel-Orderer) is designed with inductive biases in terms of both architectures and loss functions and empirically outperforms softmax models.

The benefits of Chanel-Orderer extend beyond the correction of channel order. In a typical scenario where an \texttt{RGB} image is mis-displayed in \texttt{BGR} order, Chanel-Orderer can correct the order to ensure the image is displayed correctly. This has implications for a wide range of applications, including image processing, computer graphics, and user interfaces.

Furthermore, as a byproduct of the model’s training, Chanel-Orderer also gains the ability to predict image monochromaticism (i.e. to predict whether a given image is a near-grayscale image or not). This is achieved by leveraging the model’s understanding of the semantic content of objects and their representation in color channels. Near-gray-scale images often have very similar values across all three color channels, which the model can grasp statistically and detect and classify accordingly.

\begin{figure*}[t]
    \centering
    \includegraphics[width=0.95\linewidth]{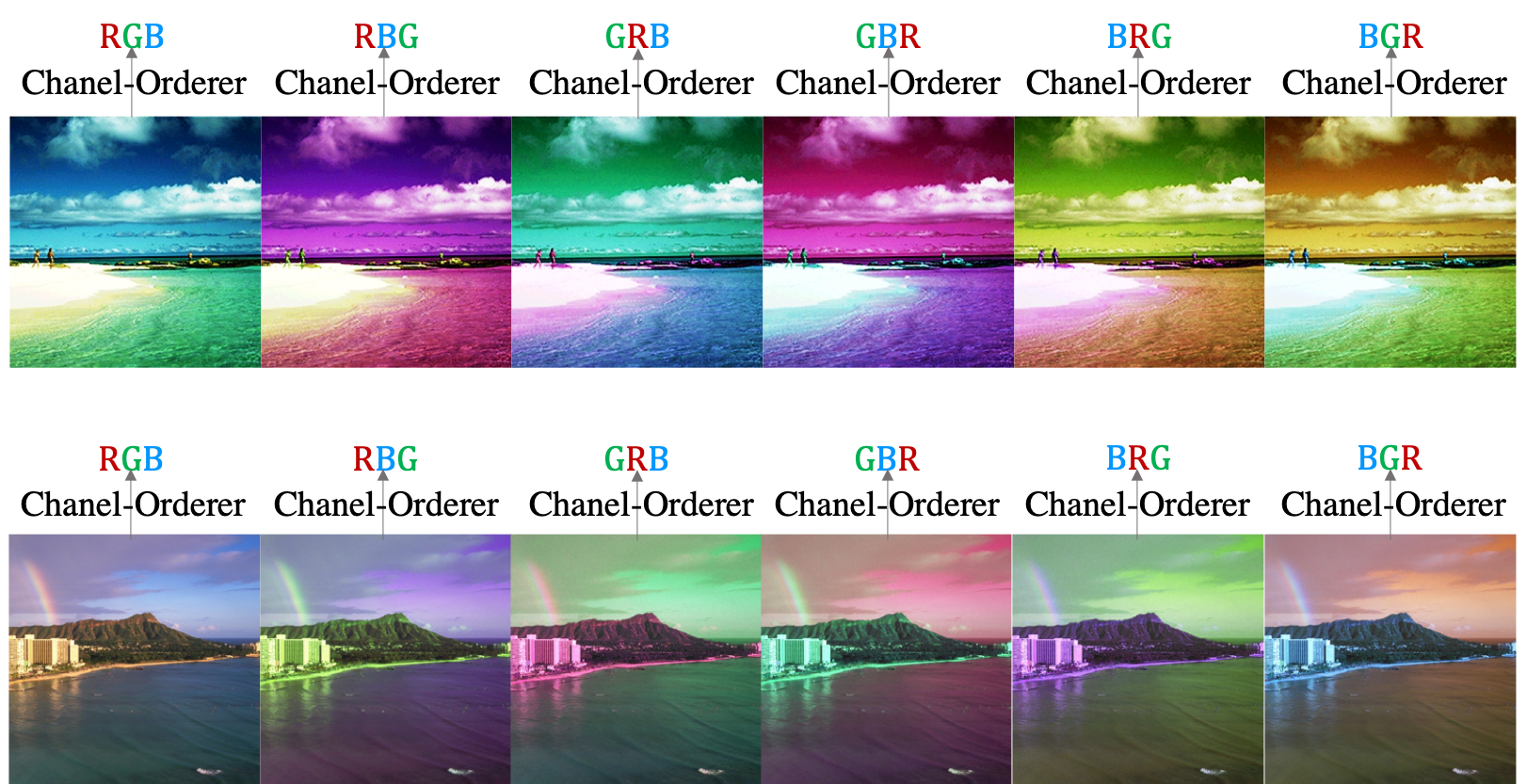}
    \caption{\small We show a proof-of-concept that, given a typical 3-channel images but in a permuted channel order, our proposed model Chanel-Orderer with ad-hoc inductive biases can accurately predict the channel ordering. Note that an alternative straightforward workaround of this problem is to cast it into a classification problem which covers $3!=6$ categories: \texttt{RGB}, \texttt{RBG}, \texttt{GRB}, \texttt{GBR}, \texttt{BRG} and \texttt{BGR} and to train a softmax classifier for predictions. However, softmax classifiers lack necessary inductive biases and are inferior to the proposed Chanel-Orderer according to our empirical findings.
    }
    \label{fig:prob_demo}
\end{figure*}  

The remainder of this paper is organized as follows. 
%Section 2 provides a brief overview of the related work in the field of image processing and the challenges associated with channel order correction. 
Section 2 details the proposed Chanel-Orderer model, including its architecture, loss functions, and the learning process. Section 3 presents the experimental setup and results, showcasing the model’s performance on various tasks, including channel order prediction and near-grayscale classification. Finally, Section 4 closes the paper by discussing limitations and potential future directions.

\section{Methodology}

We propose a channel-order predictor, Chanel-Orderer, that can predict the ordering of channels of a given 3-channel image $\mathcal{I}$ with any of 3-permutations of $\mathcal{S}:=\{R,G,B\}$, where $R, G, B$ denotes the red, green, blue channel of the image, respectively. Note that the channel ordering of an image can be determined by deciding the orderings of $\binom{3}{2}=3$ pairs of comparison: $R$ versus $G$, $R$ versus $B$ and $B$ versus $G$. We aim to design a parameterization model $f$ that can make these three pairwise decisions. We find that the design of such a model stems from two inductive biases in terms of loss function and network architecture.

% \noindent\textbf{Notations.} Throughout the rest of the the paper, we let $\mathcal{D}$ denote a real face dataset that contains face images $\mathbf{x}_0 \in \mathbb{R}^{H\times W \times 3}$. Let $\mathbf{y}$ denote a desired identity embedding and $\mathbf{s}$ be face attributes. 

\subsection{Loss Inductive Bias}

We first define the following partial order: 
\begin{equation}
\label{eq:parital_order}
    R \succ G \succ B
\end{equation}
which suggests that ideally among the three channels, the red channel $R$ should be placed in the first channel, followed by the green channel $G$ and the blue channel $B$.

Then, given a 3-channel image $\mathcal{I}$ with any of 3-permutations $\pi(\mathcal{S}):= \{I_1, I_2, I_3\}$, we formulate the model $f$ (parameterized by $\theta$) as a scoring function which outputs the ranking scores for each of the channels independently: 
\begin{equation}
    s_1 = f_{\theta}(I_1), s_2 = f_{\theta}(I_2), s_3 = f_{\theta}(I_3)
\end{equation}
These scores are interpreted as the likeness scores that \emph{should} obey the partial order \eqref{eq:parital_order}. For example, if the groundtruth suggests $I_i \succ I_j$ according to the partial order \eqref{eq:parital_order}, then we should enforce the model to output $s_i$ and $s_j$ such that $s_i > s_j$; otherwise, $s_i \le s_j$. By modifying the model to predict the probability of $s_i > s_j$:
\begin{equation}\label{eq:pij}
    p_{ij} := \mathbb{P}(s_i > s_j) = \frac{1}{1 + \exp(-g(s_i-s_j)/T)}
\end{equation}
we can formulate the ordering prediction problem into three seperate binary classification problems ($s_1$ versus $s_2$, $s_1$ versus $s_3$, $s_2$ versus $s_3$). Ideally, such a predicted probability distribution $p_{ij}$ should get close to the desired probability distribution $y_{ij}$:
\begin{equation}
    y_{ij} =
\begin{cases}
1, & \text{if }I_i \succ I_j \\
0, & \text{if }I_i \prec I_j \\
\frac{1}{2}, & \text{otherwise}\\
\end{cases}
\end{equation}
In Eq.~\eqref{eq:pij}, the scalar $T$ denotes temperature that rescales exponent to $\exp$ and the function $g$ should be an increasing differentiable function with regards to the score difference $\Delta_{ij}:=s_i - s_j$, e.g. the identity function as the simplest choice. However, we empirically find that the choice of the identity function leads to unstable optimization. In the next section, we show a better choice of $g$ that yields amenable optimization.

Formally, given any $\mathcal{I}$, we minimize the cross entropy loss between the predicted $p_{ij}$ and the groundtruth $y_{ij}$ over all the pairs of comparison (which is inherently a function of $s$ and $y$):
\begin{align}\label{eq:ce_loss}
    \notag\min_{\theta} \mathcal{L}(s, y)\\
    := \sum_{(i,j) \in \{(1,2), (1,3), (2,3)\}} -y_{ij}\log p_{ij} - (1-y_{ij})\log(1-p_{ij})
\end{align}
Plugging $p_{ij}$ and $y_{ij}$ into Eq~\eqref{eq:ce_loss} yields
\begin{align}\label{eq:ce_loss_expanded}
    \notag\min_{\theta} \mathcal{L}(s, y)\\
    \notag =  \sum_{(i,j) \in \{(1,2), (1,3), (2,3)\}} (1-y_{ij})\frac{g(s_i - s_j)}{T} \\
    + \log\left(1 + \exp\left(-\frac{g(s_i - s_j)}{T}\right)\right)
\end{align}
%Here, $\sigma$ denotes the sigmoid function.

\begin{theorem}
\label{theorem1}
    Suppose the function $g$ is a monotonically increasing differentiable function. The loss function $\mathcal{L}(s, y)$ is an increasing function with regards to the score difference $\Delta_{ij}$ when $I_i \prec I_j$ and a decreasing function with regards to $\Delta_{ij}$ when $I_i \succ I_j$, i.e.:
    \begin{equation}
        \frac{\partial{L}}{\partial{\Delta_{ij}}} =
    \begin{cases}
    >0, & \text{if }I_i \succ I_j \\
    <0, & \text{if }I_i \prec I_j \\
    %\frac{1}{2}, & \text{otherwise}\\
    \end{cases}
    \end{equation}
\end{theorem}
\begin{proof}
\begin{equation}
    \frac{\partial{L}}{\partial \Delta_{ij}} = \frac{g'(\Delta_{ij})}{T}\left((1-y_{ij}) - \frac{\exp{(-g(\Delta_{ij})/T)}}{1+\exp{(-g(\Delta_{ij})/T)}}\right)
\end{equation}
When $y_{ij}=1$, $I_i \succ I_j$ and the derivative becomes
\begin{equation}
    \frac{\partial{L}}{\partial \Delta_{ij}} = -\frac{g'(\Delta_{ij})}{T} \cdot \frac{\exp{(-g(\Delta_{ij})/T)}}{1+\exp{(-g(\Delta_{ij})/T)}} < 0
\end{equation}
When $y_{ij}=0$, $I_i \prec I_j$ and the derivative becomes
\begin{equation}
    \frac{\partial{L}}{\partial \Delta_{ij}} = \frac{g'(\Delta_{ij})}{T} \cdot \frac{1}{1+\exp{(-g(\Delta_{ij})/T)}} > 0
\end{equation}

\end{proof}
\paragraph{Remark.} 
When $y_{ij}=1$, $I_i \succ I_j$ and the loss function is a decreasing function with regard to $\Delta_{ij}$, which suggests that the minimum of $\mathcal{L}$ is attained when the score difference $\Delta_{ij}=s_i-s_j$ is largest. Hence, during training, the scoring function $f_\theta$ will adjust its learnable parameter $\theta$ to maximize the score $s_i$ and minimize the score $s_j$. When $y_{ij}=0$, $I_i \prec I_j$ and the loss function is an increasing function with regard to $\Delta_{ij}$, which suggests that the minimum of $\mathcal{L}$ is attained when the score difference $\Delta_{ij}=s_i-s_j$ is smallest. During training, the scoring function $f_\theta$ will adjust its learnable parameter $\theta$ to minimize the score $s_i$ and maximize the score $s_j$. Similar ranking spirit can be found in~\citep{burges2005learning}. Theorem~\ref{theorem1} sheds light on the design of Chanel-Orderer inference algorithm: the larger the value of $s_i$ is, the more likely $I_i$ should be placed in front among all channels ($i=1,2,3$). In Section~\ref{sec:inference}, we will show the specific algorithm design by virtue of this insight.

\subsection{Architectural Inductive Bias}
This section introduces two architectural inductive biases that are incorporated into the implementation of Chanel-Orderer: (1) the choice of $g(\cdot)$ and $T$; (2) the architectural design of the scoring function $f_{\theta}(\cdot)$.

\begin{figure*}[t]
    \centering
    \includegraphics[width=0.93\linewidth]{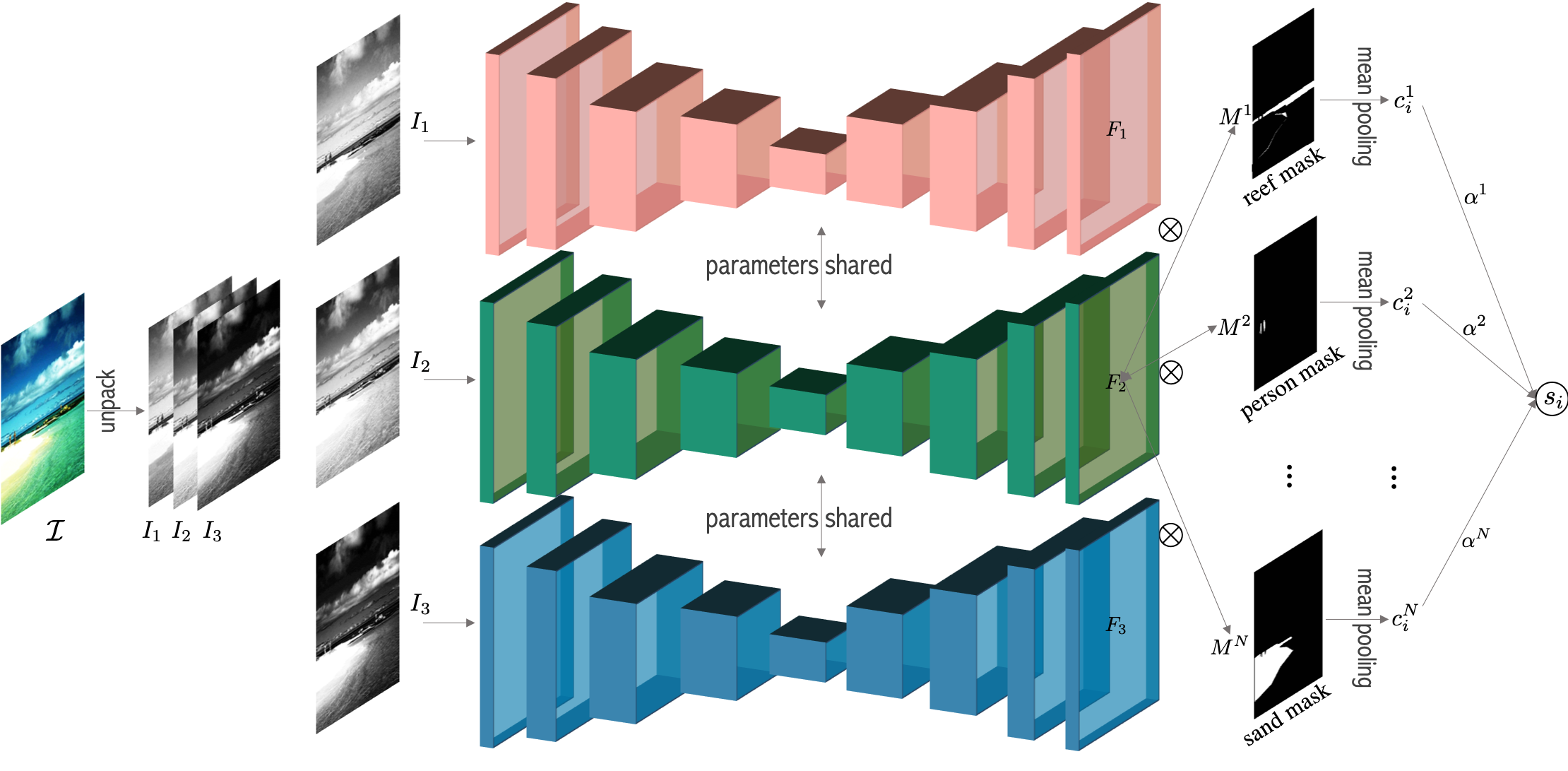}
    \caption{\small Architecture of the scoring function $f_\theta$. Given a tri-channel image $\mathcal{I}$, Chanel-Orderer first unpacks it into three channels, $I_1$, $I_2$ and $I_3$. Then, these three channels are separately and independently sent into a U-Net, which yields three feature maps $F_1$, $F_2$ and $F_3$. For each feature map $F_i$, segmentation masks $M^1, ..., M^N$ are applied to it (element-wise multiplication $\otimes$) followed by a mean pooling operation which yields the color representation for each semantic object $c_i^n$, for $n=1,...,N$. We concatenate them as a vector $c_i := [c_i^1, ..., c_i^N]^T$. The general prior weight for each object is $\alpha:=[\alpha^1, ..., \alpha^N]^T$. Then the final score $s_i$ is given by the inner product between $c_i$ and $\alpha$:, $s_i = \alpha^T c_i$.
    }
    \label{fig:architecture}
\end{figure*}

\subsubsection{Choice of $g(\cdot)$ and $T$}
As mentioned earlier, the function $g$ should be an increasing differentiable function with regard to the score difference $\Delta_{ij}$. The simplest choice is $g(\cdot)=\mathbb{I}(\cdot)$, which, however, leads to unstable optimization. We argue that this is because the distribution of $\Delta_{ij}$ does not fully overlap with the support of the sigmoid function. Here we propose another choice of $g$ that leads to amenable optimization.

According to Theorem~\ref{theorem1}, when $I_i = I_j$, the derivative $\frac{\partial{L}}{\partial{\Delta_{ij}}}$ should be zero, as no ranking should be enforced and hence no updates should be performed to the learnable parameter $\theta$. This observation suggests that $g(0)=0$:
\begin{align}
    \notag I_i = I_j \implies y_{ij}=\frac{1}{2} \\
    \notag \implies \frac{\partial{L}}{\partial \Delta_{ij}} = \frac{g'(\Delta_{ij})}{T}\left(\frac{1}{2} - \frac{\exp{(-g(\Delta_{ij})/T)}}{1+\exp{(-g(\Delta_{ij})/T)}}\right) := 0 \\
    \implies g(0) = 0
\end{align}
The last implication holds by noting that when $I_i = I_j$, the score difference $\Delta_{ij}=0$ since the scoring function $f$ is permutation-invariant. Therefore, any increasing differentiable function that passes through the origin can serve as a valid choice of $g(\cdot)$. We choose $g(\cdot):=\operatorname{tanh}(\cdot)$, as it maps $(-\infty, +\infty)$ to a symmetric domain $(-1, 1)$. To largely overlap the support of the sigmoid function, we further perform the division of $T$ which expands the range $(-1, 1)$ to the range $(-\frac{1}{T}, \frac{1}{T})$. Empirically, we set $T=0.1$ such that the resulting range $(-\frac{1}{T}, \frac{1}{T}):=(-10, 10)$ largely overlaps the definition domain of the sigmoid function, outside of which is the saturation region of the sigmoid function where gradients vanish.

\subsubsection{Architecture of $f_{\theta}(\cdot)$}
To predict the ordering of channels of a given 3-channel image, it is important to first understand the semantics of the image. Different objects in the image have different surface colors, but objects of similar semantics or of the same categories tend to exhibit similar colors in their surfaces. For example, human faces and skin, regardless of identity, tend to be yellow or brown while mountains, regardless of shape and location, tend to be green-ish. The design of the $f_{\theta}(\cdot)$ architecture should take this prior knowledge into account. Hence, the key design of our proposed Chanel-Orderer is to exploit semantic segmentation masks to predict the ranking scores.

As shown in Figure~\ref{fig:architecture}, given a three-channel image, Chanel-Orderer first separates it into three channels, $I_1$, $I_2$ and $I_3$. Then, these three channels are separately and independently sent into a U-Net~\citep{DBLP:journals/corr/RonnebergerFB15}, which yields three feature maps $F_1$, $F_2$ and $F_3$. Each feature map captures general visual representation of each image channel. For each feature map $F_i$, segmentation masks $M^1, ..., M^N$ are applied to it followed by a mean pooling operation which yields the color representation for each semantic object $c_i^n$, for $n=1,...,N$. We concatenate them as a vector $c_i := [c_i^1, ..., c_i^N]^T$. Let $\alpha:=[\alpha^1, ..., \alpha^N]^T$ denote the general prior weight for each object. Then the final score $s_i$ is given by the inner product between $c_i$ and $\alpha$:, $s_i = \alpha^T c_i$. Note that the semantic segmentation masks can be obtained from ground-truth, or from the output of a pretrained segmentation model if ground-truth is unavailable~\citep{wang2023one, wang2023internimage, su2023towards, wang2022image, fang2023eva, chen2023vision, cai2023reversible, li2023mask, jain2023semask, cheng2022masked, bousselham2021efficient, kirillov2023segment, ravi2024sam}. 
The specific training procedure is summarized in Algorithm 1.

\vspace{-1mm}
\subsection{Inference}
\label{sec:inference}
Recall that Theorem~\ref{theorem1} implies that the larger the value of $s_i$ is, the more likely $I_i$ should be placed in front among all channels ($i=1,2,3$). By virtue of this implication, we can use $s_i$ as the indicator of the channel ordering.

Specifically, given an image $\hat{\mathcal{I}}=[I_1, I_2, I_3]$ whose channels might be permuted in a wrong order, Chanel-Orderer applies its scoring function $f_\theta$ to each of the channels to obtain the scores, respectively: $s_1=f_\theta(I_1)$, $s_2=f_\theta(I_2)$, $s_3=f_\theta(I_3)$. And then label the channel with the largest score among the three as the red channel (Red), label the channel with the smallest score as the blue channel (Blue), and label the third one as the green channel (Green). See Algorithm 2 for the specific Python-like implementation.

\begin{figure*}[t]
    \centering
    \includegraphics[width=0.95\linewidth]{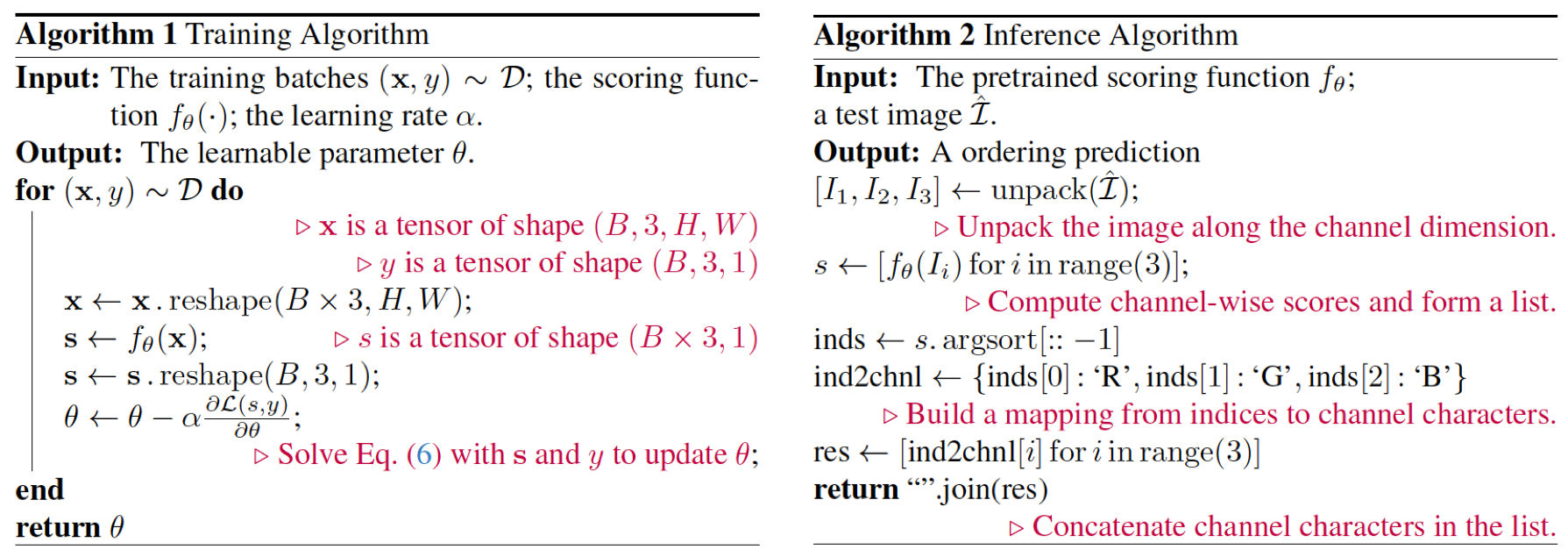}
    \label{fig:prob_demo}
\end{figure*}  

\subsection{Detection of \texttt{RGB} against \texttt{BGR}}
In most cases, we rarely encounter a scenerio where a model is expected to tell all $3!=6$ possible permutation orders. Rather, in a typical scenario, an \texttt{RGB} image is often mis-displayed in \texttt{BGR} order. To tackle this particular situation, we slightly modify the proposed Chanel-Orderer for all possible permutations into a model variant that detects \texttt{RGB} against \texttt{BGR}.

We inherit the partial order from \eqref{eq:parital_order}:
\begin{equation}
\label{eq:corollary_partial_order}
    R \succ B
\end{equation}
which suggests that ideally the red channel $R$ should be ranked ahead of the blue channel $B$ and therefore that \texttt{RGB} is preferable over \texttt{BGR}.

Given a tri-channel image $\mathcal{I}$, similarly as earlier, we first unpacks it into three channels, $I_1$, $I_2$ and $I_3$. Then, we concatenate $I_1$ and $I_2$ which yields $I_{12}$ and concatenate $I_1$ and $I_3$ which yields $I_{13}$. After a few operations followed by a global average pooling, the scoring function $f_\theta$ is expected to score $I_{12}$ and $I_{13}$ (yielding $s_{12}$ and $s_{13}$, respectively) to determine which ranks ahead of the other. To train the scoring function, a similar ranking loss function as in Eq.~\eqref{eq:ce_loss_expanded} can be applied. For inference, if $s_{12} > s_{13}$, the given image is predicted as \texttt{RGB}; otherwise, it is predicted as \texttt{BGR}.

\subsection{Detection of Near-Grayscale Images}

\begin{figure}[t]
    \centering
    \includegraphics[width=\linewidth]{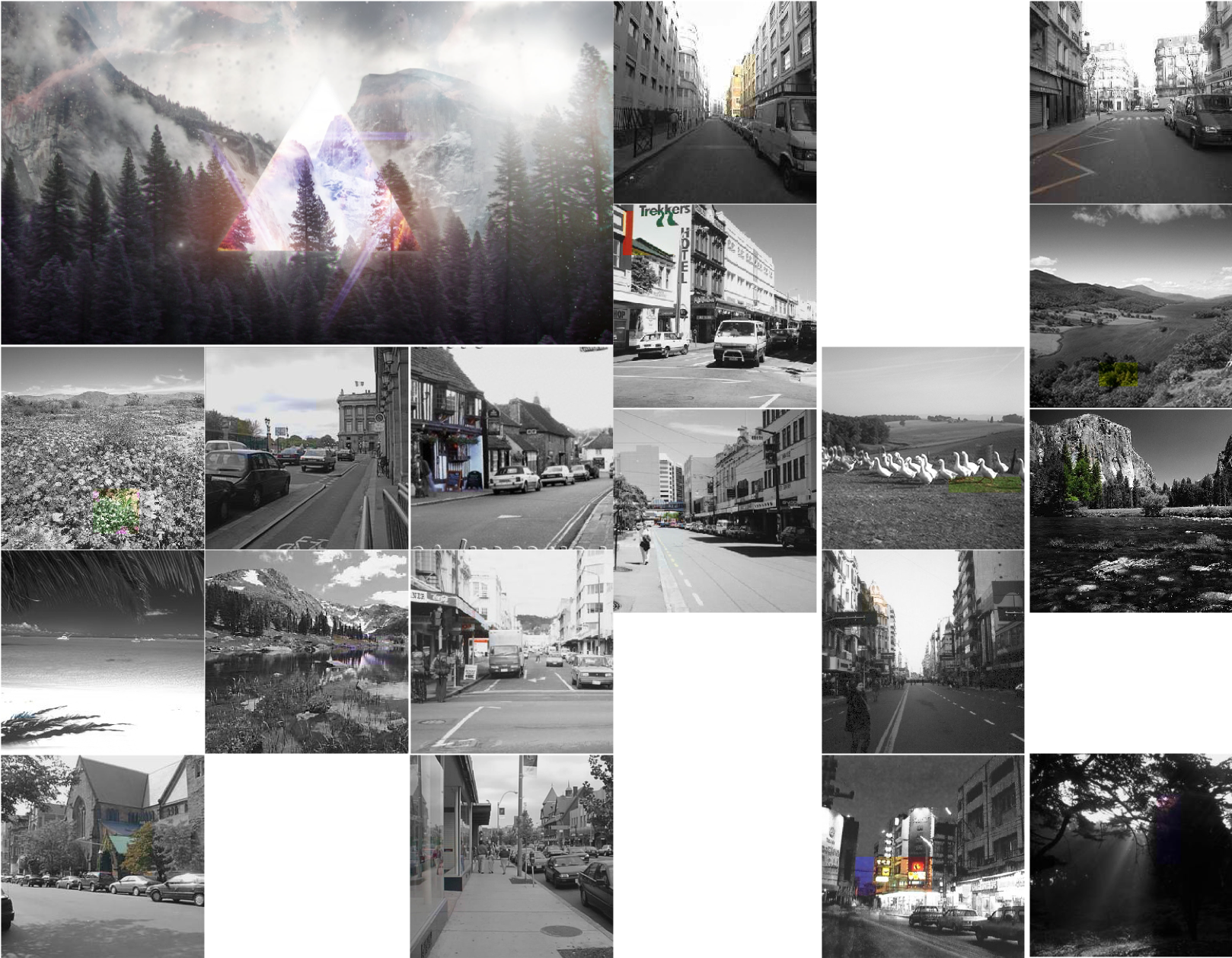}
    \caption{\small Examples of near-grayscale images. Near-grayscale images, which often appear in posters or advertisements, are mostly photographed for aesthetic purpose: photographers who make such images use polychromatic imagery to highlight the objects in the images and use monochromatic imagery to render the rest.
    }
    \label{fig:examples_of_neargray}
\end{figure}

In this section, we show our proposed Chanel-Orderer is promising in detecting near-gray images from \texttt{RGB} color images. Near-gray images are images which look monochromatic in general but have a few if not none pixels that are polychromatic (see Figure~\ref{fig:examples_of_neargray} for some examples). Such images, which often appear in posters or advertisements, are mostly photographed for aesthetic purpose: photographers who make such images use polychromatic imagery to highlight the objects in the images and use monochromatic imagery to render the rest. Prior to Chanel-Orderer, existing methods hinges upon statistic thresholding that are determined in a heuristic manner. Chanel-Orderer, in contrast, is data-driven and learns to predict the ranking scores $s_1$, $s_2$ and $s_3$ whose relative values can inherently be used as indicators to determine whether a given image is polychromatic or monochromatic. 

Specifically, given an image $\Tilde{I}$, we evaluate the ranking scores $s_i = f_\theta(\Tilde{I}_i)$, for $i=1,2,3$. And then we evaluate score differences between the three pairs which yields $\Delta_{12}$, $\Delta_{13}$, $\Delta_{23}$. Finally, we determine its monochromatism using the following rule: if $\max_{i,j}{|\Delta_{ij}|} < \tau$ (where $\tau$ is a predefined threshold), we decide it as a near-grayscale image; otherwise, it is decided as a polychromatic image.
\section{Experiments}

\subsection{Benchmarks}

We evaluate the proposed Chanel-Orderer on three challenging datasets including SiftFlow~\citep{liu2009nonparametric}, PASCAL Context~\citep{mottaghi2014role} and a customized face dataset referred to as CustoFace thereinafter. The first two benchmarks are used to evaluate the model capability on all-permutation ordering prediction, and the last one is used to evaluate the performance on the detection of \texttt{RGB} against \texttt{BGR}.

SiftFlow~\citep{liu2009nonparametric} includes 2,688 annotated images from a subset of the LabelMe database. The 256 × 256 pixel images are based on 8 different outdoor scenes, among them streets, mountains, fields, beaches, and buildings. All images belong to one of 33 semantic classes. For each test image, we permute its channels to obtain $3!=6$ versions of it. 

% Cityscapes [149] is a substantial database that concentrates on the semantic comprehension of urban street scenes. It encompasses a variety of stereo video sequences captured in urban environments across 50 cities, featuring detailed pixel-level annotations for 5,000 frames, as well as a collection of 20,000 frames with weaker annotations. The database offers semantic and dense pixel annotations across 30 classes, organized into eight categories: flat surfaces, humans, vehicles, constructions, objects, nature, sky, and void. Figure 35 illustrates four representative segmentation maps from this dataset.

PASCAL Context~\citep{mottaghi2014role} is an enhanced version of the PASCAL VOC 2010 object detection challenge, and it provides pixel-level labels for all the training images. The dataset encompasses over 400 classes (which includes the original 20 classes from PASCAL VOC, along with background classes from the segmentation dataset), categorized into three groups: objects, stuff, and hybrid categories. Due to the sparsity of many object categories in the dataset, a subset of 59 frequently occurring classes is commonly chosen for practical use.

CustoFace contains nearly 1,500 face images. All images are $128\times 128$ and contain human aligned faces across various races.

% Berkeley Segmentation Dataset (BSD) [152] contains 12,000 hand-labeled segmentations of 1,000 Corel dataset images from 30 human subjects. It aims to provide an empirical basis for research on image segmentation and boundary detection. Half of the segmentations were obtained from presenting the subject a color image and the other half from presenting a grayscale image.

We use total accuracy and accuracies in \texttt{RGB}, \texttt{RBG}, \texttt{GRB}, \texttt{GBR}, \texttt{BRG} and \texttt{BGR} to measure the model performance.

\subsection{Implementation Details}

% \begin{table*}[t]
%     \centering
%     \caption{Comparison Result on SiftFlow}
%     %\vspace{-0.1in}
%     \begin{tabular}{c|c|c|c|c|c|c|c} \toprule [1.5pt] \hline    
%             Method                                & \texttt{RGB}                   & \texttt{RBG}                       & \texttt{BGR}                          & \texttt{BRG}                   & \texttt{GBR}      & \texttt{GRB} & Overall \\ \hline
%            Shallow Model              & 46.27                 & 48.88                       & 35.82                           & 24.63                   & 27.24          &37.69  & 36.75\\ %\cline{2-9} 
%            Softmax Model         & 85.07                 & 84.70                        & 85.07                               & 84.33   & 82.46               &84.45          &84.64 \\ %\cline{2-9}
%             Chanel-Orderer-wo-Seg         & 82.46                 & 84.70                        & 83.21                          & 84.70                   & 82.09         &82.09 & 83.21 \\ %\cline{2-9}
 
%           \textbf{Chanel-Orderer}           & \textbf{85.82}      & {\color[HTML]{000000}\textbf{86.57}} & \textbf{85.82}                   & \textbf{86.57}          & \textbf{89.55} &\textbf{89.55}  & \textbf{87.31}   \\ \hline
%     \bottomrule [1.5pt]
%     \end{tabular}
%     \label{tab:siftflow}
% \end{table*}

\begin{table*}[t]
    \centering
    \caption{Comparison Result on SiftFlow}
    %\vspace{-0.1in}
    \begin{tabular}{c|c|c|c|c|c|c|c} \toprule [1.5pt] \hline    
            Method                                & \texttt{RGB}                   & \texttt{RBG}                       & \texttt{BGR}                          & \texttt{BRG}                   & \texttt{GBR}      & \texttt{GRB} & Overall \\ \hline
           Shallow Model              & 46.27                 & 48.88                       & 35.82                           & 24.63                   & 27.24          &37.69  & 36.75\\ %\cline{2-9} 
           Softmax Model         & 85.07                 & 84.70                        & 85.07                               & 84.33   & 82.46               &84.45          &84.64 \\ %\cline{2-9}
            Chanel-Orderer-wo-Seg         & 82.46                 & 84.70                        & 83.21                          & 84.70                   & 82.09         &82.09 & 83.21 \\ %\cline{2-9}
 
          \textbf{Chanel-Orderer}           & \textbf{98.51}      & {\color[HTML]{000000}\textbf{98.51}} & \textbf{98.51}                   & \textbf{98.51}          & \textbf{98.51} &\textbf{98.51}  & \textbf{98.51}   \\ \hline
    \bottomrule [1.5pt]
    \end{tabular}
    \label{tab:siftflow}
\end{table*}

\begin{table*}[t]
    \centering
    \caption{Comparison Result on PASCAL-Context}
    %\vspace{-0.1in}
    \begin{tabular}{c|c|c|c|c|c|c|c} \toprule [1.5pt] \hline    
            Method                                & \texttt{RGB}                   & \texttt{RBG}                       & \texttt{BGR}                          & \texttt{BRG}                   & \texttt{GBR}      & \texttt{GRB} & Overall \\ \hline
           Shallow Model              & 30.30                 & 30.50                       & 38.02                           & 40.00                   & 34.65          &35.64 & 34.85 \\ %\cline{2-9} 
           Softmax Model         & \textbf{77.42}                 & 74.06                        & 75.25                               & 74.06   & 67.52               &71.68          &73.33 \\ %\cline{2-9}
            Chanel-Orderer-wo-Seg         & 57.43                 & 57.82                        & 60.40                          & 59.01                   & 58.42        &57.62  & 58.45 \\ %\cline{2-9}
 
          \textbf{Chanel-Orderer}           & 73.86      & \textbf{74.46}                   & \textbf{78.22}          & \textbf{79.60} &\textbf{74.26}  & \textbf{74.06}  & {\color[HTML]{000000}\textbf{75.74}} \\ \hline
    \bottomrule [1.5pt]
    \end{tabular}
    \label{tab:pascal}
\end{table*}

The proposed Chanel-Orderer consists of a U-Net architecture~\citep{DBLP:journals/corr/RonnebergerFB15} with the four layers of encoders that maps an input into 32-channel, 64-channel, 128-channel and 256-channel sequentially, then with a four layers of decoders that map the encoded feature map back to 128-channel, 64-channel, 32-channel and 1-channel. The intermediate activation functions are ReLUs. The training batch size is set to $48$ and the total training epochs is $100$. The initial learning rate is set to $0.001$ and decays with the factor of $0.98$
Throughout the entire training process, we use the Adam optimizer.

\subsection{Performance Evaluation}

\subsubsection{Competing Methods}
We compare our proposed Chanel-Orderer with other promising methods, including shallow models, Softmax models and other Chanel-Orderer variants.

\paragraph{Shallow models:} we construct color histograms~\citep{novak1992anatomy} for each channel of images $\mathbf{h}_1$, $\mathbf{h}_2$ and $\mathbf{h}_3$, and train a simple classifier $F$ to tell which should come first given a pair of channels. That is, for each $(i,j) \in \{(1, 2), (1, 3), (2, 3)\}$, train the classifier $F$ to take as input the concatenated color histograms $[\mathbf{h}_i, \mathbf{h}_j]$ and output the probability that the $i$-th channel ranks in the front of the $j$-th channel according to the predefined partial order shown in Eq.~\eqref{eq:parital_order}.
  
\paragraph{Softmax models~\citep{bridle1989training}:} in this model, we formulate the ordering prediction task as a multi-class classification task, that is, to train a classifier to predict which category a given image should fall into: $\texttt{RGB}$, $\texttt{RBG}$, $\texttt{GRB}$, $\texttt{GBR}$, $\texttt{BRG}$ and $\texttt{BGR}$. For the detection of $\texttt{RGB}$ against $\texttt{BGR}$, the classifier is to predict $\texttt{RGB}$ or $\texttt{BGR}$ only. For the detection of near-grayscale images, as the classifer outputs a categorical distribution over all $3!=6$ categories, we use its entropy as an indicator of monochromatism (see the next section for the specifics).
  
\paragraph{Chanel-Orderer-wo-Seg:} our proposed Chanel-Orderer exploits the segmentation semantics to help make the ordering predictions. To investigate the effect of segmentation semantics, we perform an ablation study by removing the segmentation semantics. Specifically, we remove the element-wise multiplication between $F_i$ and $M^n$ and only leave the mean pooling operation upon $F_i$. The resulting model is referred to as Chanel-Orderer-wo-Seg. We compare Chanel-Orderer against it for the ablation study on the effect of segmentation semantics.

% \subsubsection{Qualitative Results}
% Here, we illustrate a collection of face images generated by \propmod as qualitative evaluation.
% \figureautorefname~\ref{fig:generated_faces} shows the results for randomly sampled identities (IDs) under various attribute conditions;
% %such as age, pose, illumination, facial expression, with glasses, and hairstyle;
% %each ID under a specific condition includes 7 sample cases to show the diversity in generation.
% Obviously, when comparing different identities (inter-class), 
% the essential intrinsic key information of each identity is still retained and can be easily identified. Also, different samples of each identity (intra-class) exhibit distinct diversity, stemming from variations in similarity scores ($\nu_{ij}$'s) and differences in face attributes as conditioning signals.
% In terms of the effect of our proposed adjusted score and the original score on the sampling algorithm, we observe that the face images generated by our proposed \propmod exhibits much better quality and identity preservation than those generated by the original score function, as shown in Figure~2.
%\ref{fig:grad_show}.

% \begin{figure}[t]
%     \centering
%     \includegraphics[width=\linewidth]{Fig/Presentation3.pdf}
%     %\vspace{-0.2in}
%     \caption{Qualitative comparison of face images generated by the adjusted score function $\nabla \log \tilde{p}(\mathbf{x}_t|\mathbf{y}, \mathbf{s})$ and the original score function $\nabla \log p(\mathbf{x}_t|\mathbf{y}, \mathbf{s})$.}
%     \label{fig:grad_show}
% \end{figure}

\subsubsection{Quantitative Results}

The comparison results on SiftFlow are shown in Table~\ref{tab:siftflow}. The Chanel-Orderer model achieves the best overall performance with the overal accuracy of 98.51\%. It is the most robust model to changes in channel order since it maintains high accuracies across all channel orders. The Softmax Model also performs well with an overall average of 84.64\%, indicating that it is less sensitive to channel order than the Shallow Model, which shows significant drops in performance with certain channel orders. The Chanel-Orderer-wo-Seg model performs similarly to the “Softmax Model” but slightly less robustly to channel order changes. Shallow Model has a wide range of performance scores, indicating high sensitivity to the input channel order. The highest accuracy is 48.88\% for the \texttt{RGB} channel order, and the lowest is 24.63\% for the \texttt{BRG} channel order. The overall average accuracy is 36.75\%, which is the lowest among the models tested.
Softmax Model performs significantly better than the Shallow Model, with a high degree of consistency across different channel orders. The overall average accuracy is 84.64\%, with the lowest accuracy being 82.46\% for the \texttt{GBR} channel order.
Chanel-Orderer-wo-Seg also performs well, with an overall average accuracy of 83.21\%. The performance is quite consistent, with the accuracy ranging from 82.09\% to 84.70\%. This suggests that the model is less sensitive to channel order changes compared to the Shallow Model.
Chanel-Orderer has the highest overall average accuracy at 98.51\%. It shows a very consistent performance across all channel orders, with the lowest accuracy being 98.51\% and the highest being 98.51\%. This indicates that the Chanel-Orderer model is highly robust to variations in channel order.

The comparison results on PASCAL-Context are shown in Table~\ref{tab:pascal}.  Shallow Model has a varied performance across different channel orders, with the highest accuracy of 40.00\% for the \texttt{BRG} channel order and the lowest of 30.30\% for the \texttt{RGB} channel order. The overall average accuracy is 34.85\%, which is the lowest among the models tested. This suggests that the Shallow Model is not only performing poorly overall but is also highly sensitive to the input channel order.
Softmax Model shows better performance than the Shallow Model across all channel orders, with an average accuracy of 73.33\%. The performance is relatively consistent, except for a noticeable drop when the channel order is \texttt{GBR}, where the accuracy drops to 67.52\%. This indicates that while the Softmax Model is more robust to channel order changes than the Shallow Model, it is still somewhat affected by them.
Chanel-Orderer-wo-Seg has an overall average accuracy of 58.45\%, which is lower than the Softmax Model but higher than the Shallow Model. The performance is relatively stable across different channel orders, with a narrow range from 57.43\% to 60.40\%. This suggests that the model is designed to handle channel order variations to some extent, but it is not as effective as the Chanel-Orderer model.
Chanel-Orderer has the highest overall average accuracy at 75.74\%, which is significantly better than the other models. It also shows the most consistent performance across different channel orders, with a narrow range from 73.86\% to 79.60\%. This indicates that the Chanel-Orderer model is highly effective at dealing with channel order variations and is the most robust model in this comparison.

%\vspace{-2mm}
\paragraph{Detection of \texttt{BGR} against \texttt{RGB}.}
We compare Chanel-Orderer with the Softmax model. As shown in Table~\ref{tab:bgr_rbg}, Chanel-Orderer achieves the accuracy of 93.85\% whereas the Softmax model only achieves 51.63\%. This suggests that without sufficient inductive biases either in terms of architecture or loss, the Softmax model is unable to take any shortcut to learn a valid mapping for classification. Chanel-Orderer, however, casts this problem as a ranking problem and makes use of the architectural and loss inductive biases to learn the ranking, and therefore achieves promising results on this task.

\paragraph{Detection of Near-Grayscale Images.}
We compare Chanel-Orderer against the Softmax model in the detection of near-gray images. Recall that Chanel-Orderer uses the maximum absolute score difference $\max_{i,j}{|\Delta_{ij}|}$ as an indictor to detect near-grayscale images. If $\max_{i,j}{|\Delta_{ij}|} \le \tau$ ($\tau$ is a predefined threshold), the given image is detected as near-grayscale; otherwise, it is detected as RGB. On the other hand, the Softmax model outputs $3!=6$ probabilities ($p_i$ for $i=1,...,6$) for each color orderings. We use the softmax entropy as the indictor of monochromatism:
\begin{equation}
    H[p] = -\sum_{i=1}^6 p_i \log p_i
\end{equation}
since if the softmax entropy is high, the softmax model has high epistemic uncertainty~\citep{Xu_2023_CVPR} about the channel ordering of a given image.

\begin{figure}[t]
    \centering
    \includegraphics[width=\linewidth]{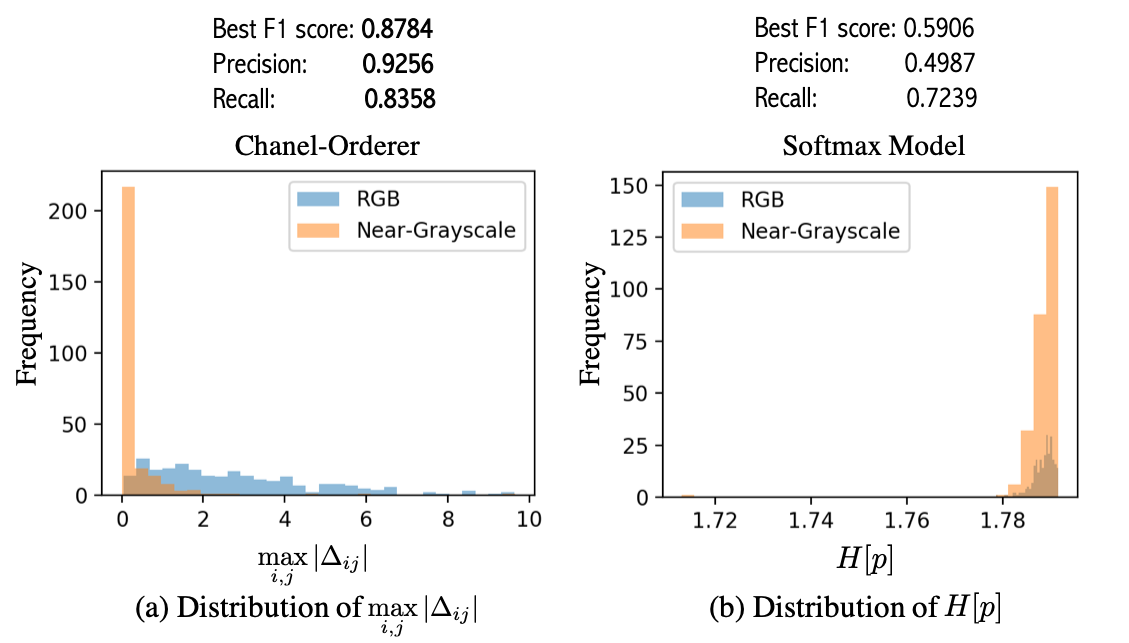}
    \caption{\small Detection of near-grayscale images. (a) Results of Chanel-Orderer and the distribution of $\max_{i,j}{|\Delta_{ij}|}$. The threshold $\tau$ is set to $0.4$. (b) Results of Softmax Model and the distribution of $H[p]$. The threshold is set to $1.79$.
    }
    \label{fig:detection_of_neargray}
\end{figure}

As shown in Figure~\ref{fig:detection_of_neargray}, we observe that Chanel-Orderer outperforms the Softmax model by clear margins in this task: the maximum absolute score difference $\max_{i,j}{|\Delta_{ij}|}$ given by Chanel-Orderer can distinguish near-grayscale images from normal RGB images whereas the entropy $H[p]$ given by Softmax model cannot. Consequently, Chanel-Orderer achieves F1-score of 0.8784 while Softmax model only achieves 0.5906.
According to prior works~\citep{guo2017calibration, gal2016dropout, pearce2021understanding} on softmax, neural networks trained by softmax loss tend to yield miscalibrated probabilities on the basis of information that is not meant for desired predictions to human intelligence. 

\subsection{Model Behavoir Analysis}
The results from Table~\ref{tab:siftflow}, Table~\ref{tab:pascal} and Table~\ref{tab:bgr_rbg} suggest that Chanel-Orderer consistently outperforms Softmax models in almost all cases. Softmax models cast the channel-ordering prediction into a classification problem whereas Chanel-Orderer tackles this problem in ranking spirit. This further suggests that ranking is more preferable as inductive bias than classification in this particular task. This can also be seen from the training progress: we observe, during training, that Chanel-Orderer converges much faster than Softmax models into smaller loss values, which validates the advantage of inductive biases incorporated into the model.
\section{Conclusion}
The advent of digital imaging has revolutionized our ability to capture, store, and process visual information, yet it has also introduced complexities such as the correct interpretation of image data. This paper presents Chanel-Orderer, a statistical ranking model designed to address the challenge of determining the correct channel order of color images, a task that is pivotal for accurate image representation and subsequent analysis. Through our proof-of-concept, we have demonstrated the model’s capability to accurately predict the original channel order of images, even when the channels are permuted, thereby mitigating issues related to incorrect display or processing.

Our approach, which leverages ad-hoc inductive biases in terms of loss function and architecture, has proven to be effective in scoring each color channel based on these semantic priors. Chanel-Orderer not only ensures the correct display of image channels but also extends its utility to predicting image monochromatism in a statistical prospective.

The implications of Chanel-Orderer’s success are far-reaching, touching upon various domains including image processing, computer graphics, and user interface design. By ensuring images are accurately represented, Chanel-Orderer contributes to an enhanced user experience, more reliable processing outcomes, and increased efficiency in the development of imaging applications.

Looking forward, there are several avenues for future research. First, we aim to generalize the model to accommodate a broader range of color spaces and channel configurations, expanding its applicability. Second, integrating Chanel-Orderer with existing imaging libraries and software ecosystems will be a key step towards streamlining image handling across diverse platforms. Finally, we are committed to improving the model’s robustness and accuracy to cater to the vast array of image conditions encountered in real-world scenarios.

% In conclusion, the Chanel-Orderer model represents a significant stride in the advancement of digital imaging technologies. It underscores the potential of machine learning to resolve intricate challenges in image representation and processing, setting the stage for further innovations in the field.

% \noindent{\textbf{Limitations.}}
% While $\text{ID}^3$, designed for the sake of privacy protection, achieves SoTA performance in SFR, there remains clear margins as compared to the FR performance when training with real-world face datasets such as MS1M. This suggests that the fake face dataset generated by $\text{ID}^3$ does not fully approximate the real-world faces. Future work might include closing this gap.

\begin{table}[t]
    \centering
    \caption{Detection of \texttt{BGR} against \texttt{RGB}}
    %\vspace{-0.1in}
    \begin{tabular}{c|c} \toprule [1.5pt] \hline    
            Method                                & Accuracy                   \\ \hline
           Softmax Model         & 51.63 \\ %\cline{2-9}
          \textbf{Chanel-Orderer}           & \textbf{93.85}  \\ \hline
    \bottomrule [1.5pt]
    \end{tabular}
    \label{tab:bgr_rbg}
\end{table}

\paragraph{Limitations.}
While the Chanel-Orderer model has shown promise in addressing the challenge of correcting color channel order, it is essential to acknowledge its potential limitations. These limitations provide insights into areas for further research and development.

- Generalization: The model’s performance may be limited to specific types of images or datasets. As the model’s inductive biases are tailored to learn object semantics, it may struggle with images that include open-set semantic categories. Expanding the model’s training data and exploring more diverse image categories could enhance its generalization capabilities.

- Complexity: The complexity of the model’s architecture and the need for specialized training data may pose challenges for deployment in resource-constrained environments. Simplifying the model or developing lightweight versions could make it more accessible for a wider range of applications.

- Sensitivity to Image Quality: The model’s performance may be sensitive to the quality of the input images. Issues such as noise, compression artifacts, or pixelation may hinder its ability to accurately predict the original channel order. Improving the model’s robustness to such challenges is a critical area for future work.

% - Limited Channel Configurations: The model is currently designed to handle the commonly used tri-channel color spaces. Extending its capabilities to other color spaces and channel configurations would broaden its applicability and utility in diverse imaging scenarios.

Future work might focus on addressing these challenges for better performance. %By doing so, we aim to develop a robust and versatile tool that can be seamlessly integrated into existing imaging workflows and systems, ultimately enhancing the quality and reliability of digital imaging technologies.

{
    \small
    \bibliographystyle{ieeenat_fullname}
    \bibliography{main}
}


\end{document}